\documentclass[11pt,a4paper]{article}
\PassOptionsToPackage{breaklinks}{hyperref}
\usepackage{xurl}
\usepackage[hyperref]{acl2021}
\usepackage{times}
\usepackage{latexsym}
\usepackage{booktabs}

\usepackage{soul}

\usepackage{microtype}

\aclfinalcopy % Uncomment this line for the final submission
\def\aclpaperid{22} %  Enter the acl Paper ID here

\title{Between Flexibility and Consistency: \\
Joint Generation of Captions and Subtitles}

\author{Alina Karakanta$^{1,2}$, Marco Gaido$^{1,2}$, Matteo Negri$^1$, Marco Turchi$^1$ \\
$^1$ Fondazione Bruno Kessler, Via Sommarive 18, Povo, Trento - Italy  \\ $^2$ University of Trento, Italy\\
	{\texttt{\{akarakanta,mgaido,negri,turchi\}@fbk.eu}}}

\date{}

\begin{document}
\maketitle
\begin{abstract} %200 words
Speech translation (ST) has lately received growing interest for the generation of subtitles without the need for an intermediate source language transcription and timing (i.e. captions). However, the joint generation of source captions and target subtitles does not only bring potential output quality advantages when the two decoding processes inform each other, but it is also often required in multilingual scenarios. In this work, we focus on ST models which generate consistent captions-subtitles in terms of structure and lexical content. We further introduce new metrics for evaluating subtitling consistency. Our findings show that joint decoding leads to increased performance and consistency between the generated captions and subtitles while still allowing for sufficient flexibility to produce subtitles conforming to language-specific needs and norms. 
\end{abstract}

\section{Introduction}
New trends in media localisation call for the rapid generation of subtitles for vast amounts of audiovisual content. 
Speech translation, and especially direct approaches \cite{berard_2016,bahar2019comparative}, have recently shown promising results with high efficiency because they do not require a transcription (manual or automatic) of the source speech but generate the target language subtitles directly from the audio. 
However, obtaining the intralingual subtitles (hereafter  ``captions'') is necessary for a range of applications, while in some settings captions need to be displayed along with the target language subtitles. Such ``bilingual subtitles'' are useful in multilingual online meetings, in countries with multiple official languages, or for language learners and audiences with different accessibility needs. In those cases, captions and subtitles should not only be consistent with the visual and acoustic dimension of the audiovisual material but also between each other, for example in the number of blocks (pieces of time-aligned text) %\mn{(i.e. the portions of text appearing altogether on screen)} 
they occupy, their length and segmentation. Consistency is vital for user experience, for example in order to elicit the same reaction among multilingual audiences, or %and ensure that no one is left behind. Moreover, consistency between captions and subtitles is important in the localisation industry, 
to facilitate the quality assurance process in the localisation industry. %and to reduce the effort of post-processing each language version separately. %since errors in the format or lexical content in the captions should also be fixed in the subtitles.

Previous work in ST for subtitling has focused on 
%the target language subtitles
generating interlingual subtitles
\cite{matusov-etal-2019-customizing,karakanta-etal-2020-42}, a) without considering the necessity of obtaining captions consistent with the target subtitles, and b) without examining whether the joint generation leads to improvements in quality. 
%\mn{[I liked the passage above in the .tex, which now is commented. See the comment.]}
%However, the joint generation of captions and subtitles can have several benefits.
% Firstly, joint generation with a single system can increase efficiency by 
% %preventing
% \mn{avoiding}
% the maintenance of two different models and therefore speed up the localisation process.
We hypothesise that knowledge sharing between the tasks of transcription and translation could lead to such improvements. Moreover, joint generation with a single system can avoid the maintenance of two different models, increase efficiency, and in turn speed up the localisation process. 
%Firstly, joint generation with a single system can avoid the maintenance of two different models, increase efficiency, and in turn speed up the localisation process. Second, we hypothesise that %joint generation
%it could lead to improvements in quality, due to the knowledge sharing between the tasks of transcription and translation.
Lastly, if joint generation improves consistency, joint models could increase automation in subtitling applications where consistency is a desideratum.

In this work, we address these issues for the first time, by jointly generating both captions and subtitles 
%from the 
for the same audio source. We experiment with the following models: 1) \textbf{Shared Direct} \cite{weiss2017sequence}, where the speech encoder is shared between the transcription and the translation decoder, %and 2) \textbf{Triangle} \cite{anastasopoulos-chiang-2018-tied}, where transcript decoder states are passed to the translation decoder. % 
2) \textbf{Two-Stage} \cite{kano-2017-2stage}, where the transcript decoder states are passed to the translation decoder, and 3) \textbf{Triangle} \cite{anastasopoulos-chiang-2018-tied}, which extends the two-stage by adding a second attention mechanism to the translation decoder which attends to encoded speech inputs. 
We compare these models with the established approaches in ST for subtitling: an independent direct ST model and a cascade (ASR+MT) model. %used in previous work. 
Moreover, we extend the evaluation 
%of our models 
beyond 
% transcription (WER) and translation (BLEU) quality, by evaluating the form of the generated subtitles and subtitling consistency. 
the usual metrics used to assess transcription and translation quality (respectively WER and BLEU), by also evaluating the form and consistency of the generated subtitles.

\citet{sperber-etal-2020-consistent} introduced several lexical and surface  metrics to measure consistency of ST outputs, but they were only applied to standard, non-subtitle, texts.
Subtitles, however, are a particular type of text %\mn{, which is} 
structured in blocks which accompany the action on screen. Therefore, we propose to measure their consistency by taking advantage of this structure and introduce metrics able to reward subtitles that share similar structure and content. 

Our contributions can be summarised as follows:
\begin{itemize}
    \item We employ ST to directly generate both captions and subtitles without the need for human pre-processing (transcription, segmentation).
    \item We propose new, task-specific metrics to evaluate %the form of subtitles and 
    subtitling consistency, a challenging and understudied problem in subtitling evaluation.
    \item We show increased performance and consistency between the generated captions and subtitles compared to independent decoding, while preserving adequate conformity to subtitling constraints.

\end{itemize}

\section{Background} \label{sec:related}
\subsection{Bilingual subtitles}
New life conditions maximised the time spent in front of screens, transforming today's mediascape in a complex puzzle of new actors, voices and workflows. Face-to-face meetings and conferences moved online, opening up participation for global audiences. In these new settings multilinguality and versatility are dominant, and manifested in business meetings with international partners, conferences with world-wide coverage% with the participation of populations with different dominant languages
, multilingual classrooms and audiences with mixed accessibility needs. %inclusion. All these together with the tendency for customisation, personalisation and personal preferences on experiencing audiovisual material urge us to find ways in which technology can support these growing needs. 
%In a world that becomes increasingly multilingual and given the abundance of audiovisual material produced for entertainment, education and communication, %subtitles are becoming more and more essential and serve a variety of functions. 
Given these growing needs for providing inclusive and equal access to audiovisual material for a multifaceted audience spectrum, 
efficiently obtaining high-quality captions and subtitles is becoming increasingly relevant.
% in many scenarios.

%\mt{I have never thought about generating consistent subtitling in two languages for the scenarios you presented here. It is interesting! My initial thought was about generating both of them with a single system to reduce the maintenance of two systems and to have speed up the process. }
Traditionally, displaying subtitles in two languages in parallel (bilingual or dual subtitles) has been common in countries with more than one official languages, such as Belgium and Finland \cite{gottlieb-2004}. %In this case, both language versions displayed are usually target language versions and rarely captions. 
Recently, however, captions along with subtitles have been employed in other countries to attract wider audiences, e.g. in Mainland China English captions are displayed along with Mandarin subtitles. Interestingly, despite doubling the amount of text that appears on the screen and the high redundancy, it has been shown that bilingual subtitles do not significantly increase 
%the
users' cognitive load %involved in their comprehension 
\cite{Liao2020TheIO}.

One group which undoubtedly benefits from the parallel presence of captions and subtitles are language learners. %The value of audiovisual materials has been recognised both for foreign (L2) language learning and in education, with the proliferation of Massive Open Online Courses (MOOCs). 
Captions have been found to increase learners' L2 vocabulary \cite{Sydorenko-2012-vocab} and improve listening comprehension \cite{guichon-2008-learn}. Subtitles in the learners' native language (L1) are an indispensable tool for comprehension and access to information, especially for beginners. 
In bilingual subtitles, the captions support learners in understanding the speech and acquiring terminology% in their L2
, while subtitles serve as a dictionary, facilitating bilingual mapping \cite{garcia-2017-bisubsf}. % found that bilingual subtitles in short educative videos help learners acquire L2 vocabulary and better understand and recall concepts of the subject matter. 
Consistency is particularly important for bilingual subtitles. Terms should fall in the same block and in similar positions. Moreover, similar length and equal number of lines %in a parallel subtitle block
can prevent long distance saccades, assisting in spotting the necessary information in the two language versions. Several subtitling tools have recently allowed for combining captions and subtitles on the same video (e.g. Dualsub\footnote{\url{https://www.dualsub.xyz/}}) and bilingual subtitles can be obtained for Youtube videos\footnote{\url{https://www.watch-listen-read.com/}} and TED Talks.\footnote{\url{https://amara.org/en/teams/ted/}}

Another aspect where consistency between captions and subtitles is present is in subtitling templates. A subtitling template is a source language/English version of a subtitle file already segmented and containing timestamps, which is used to directly translate the text in target languages while preserving the same structure \cite{Avt2007sub,georgakopoulou-19,netflix-template}. This process reduces the cost, turn-around times and effort needed to create a separate timed version for each language, and %, and opens up the pool of translators able to perform the task. Another benefit is that it 
facilitates quality assurance since errors can be spotted across the same blocks \cite{Nikolic2015}. These benefits motivated our work towards simultaneously generating two language versions with the maximum consistency, where the caption file can further serve as a template for multilingual localisation. 
% %Our work 
 This paper is a first step towards maximising automation for the generation of high-quality multiple language/accessibility subtitle versions.
%\mn{This paper is a first step along this direction.}

\subsection{MT and ST for subtitling}
Subtitling has long sparked the interest of the Machine Translation (MT) community as a challenging type of translation. Most works employing MT for subtitling stem from the statistical era \cite{Volk-et-al-2010,Etchegoyhen-2014-sumat} %,aziz-etal_EAMT:2012}
or even before, with example-based approaches \cite{melero-2006-etitle,armstrong-2006-dvd,nyberg-1997-captions,Popowich-2000-mtcaptiona,piperidis-2005-musa}.
With the neural era, the interest in automatic approaches to subtitling revived. Neural Machine Translation (NMT) led to higher performance and efficiency and opened new paths and opportunities. \newcite{matusov-etal-2019-customizing} customised a cascade of ASR and NMT for subtitling, using domain adaptation with fine-tuning and improving subtitle segmentation with a specific segmentation module. Similarly, using cascades, \citet{koponen-2020} explored sentence- and document-level NMT for subtitling and showed productivity gains for some language pairs. %Among the most promising approaches, however, are end-to-end systems, which do not rely on the presence of separately trained components for ASR, MT and segmentation \cite{karakanta-etal-2020-42}.
However, bypassing the need to train and maintain separate components for transcription, translation and segmentation, direct end-to-end ST systems are now being considered as a valid and potentially more promising alternative \cite{karakanta-etal-2020-42}. Indeed, besides the architectural advantages, they come with the promise to avoid error propagation (a well known issue of cascade solutions), reduce latency, and better exploit speech information (e.g. prosody) without loss of information thanks to a less mediated access to the source utterance.
To our knowledge,
%to date,
no previous work has yet explored the effectiveness of joint automatic generation of captions and subtitles.

\subsection{Joint generation of transcription and translation}
The idea of generating transcript and translation has been previously addressed in \cite{weiss2017sequence,anastasopoulos-chiang-2018-tied}. These papers presented different solutions (e.g. shared decoder and triangle) with the goal of improving translation performance by leveraging both ASR and ST data in direct ST. Later, \citet{sperber-etal-2020-consistent} evaluated these methods with the focus of jointly producing consistent source and target language texts. 
Their underlying intuition is that, since in cascade solutions the translation is derived from the transcript, cascades should achieve higher consistency than direct solutions. Their results, however, showed that triangle models achieve the highest consistency among the architectures tested (considerably better than that of cascade systems) and have competitive performance in terms of translation quality. Direct independent and shared models, instead, do not achieve the translation quality and consistency of cascades. However, all these previous efforts fall outside the domain of automatic subtitling and ignore the inner structure of the subtitles and their relevance when considering consistency.

%\st{The models we are testing in the present work have been examined in previous work for the joint generation of transcription and translation} \cite{sperber-etal-2020-consistent}. \st{However, these previous efforts fall outside the domain of automatic subtitling.} %Our selection of models for this work is narrower than in . This decision is motivated by the fact that we have selected the most competitive model from each category (DirMu for the direct model category, Triangle for the joint model category) and that models inside each category perform similarly.  
%\st{Since in cascade solutions the translation is derived from the transcript, they intuitively should be more consistent than outputs of direct solutions.
%In contrast, \citet{sperber-etal-2020-consistent} showed that triangle models achieve the highest consistency among the architectures tested (considerably better than that of cascade systems) and have competitive performance in terms of translation quality. Direct independent and shared models, instead, do not achieve the translation quality and consistency of cascade solutions.

\section{Methodology}

\subsection{Models}
\label{subsec:models}
%\todo{MG} %% Expectations, how are they connected to consistency/performance?
%We experiment with the following models:
To study the effectiveness of the different existing ST approaches in the subtitling scenario, we experiment with the following models:

The \textbf{Multitask Direct Model (DirMu)} model consists of a single audio encoder and two separate decoders \cite{weiss2017sequence}: one for generating the source language captions, and the other for the target language subtitles. The weights of the encoder are shared. The model can exploit knowledge sharing between the two tasks, but allows for some degree of flexibility since inference for one task is not directly influenced by the other task.

The \textbf{Two-Stage (2ST)} model \cite{kano-2017-2stage} also has two decoders, but the transcription decoder states are passed to the translation decoder. %the translation decoder attending to the transcription decoder output embeddings, but 
This is the only source of information for the translation decoder as it does not attend to the encoder output.

The \textbf{Triangle (Tri)} model  \cite{anastasopoulos-chiang-2018-tied} is similar to the two-stage model, but with the addition of an attention mechanism to the translation decoder, which attends to the output embeddings of the encoder. Both \texttt{2ST} and \texttt{Tri} support coupled inference and joint training.

We compare these models with common solutions for ST. The \textbf{Cascade  (Cas)} model is a combination of an ASR + NMT components; the ASR transcribes the audio into text in the source language, which is then passed to an NMT system for translation into the target language. The two components are trained separately and can therefore take advantage of richer data for the two tasks. The cascade features full dependence between transcription and translation, which will potentially lead to high consistency.

The \textbf{Direct Independent (DirInd)} system consists of two independent direct ST models, one for the transcription (as in the ASR component of the cascade) and one for the translation. It hence lies on the flexibility edge of flexibility-consistency spectrum compared to the models above.

\subsection{Evaluation of subtitling consistency}\label{sec:consistency-def}
%The specific format and style requirements of subtitling call for extending the evaluation beyond transcription (WER) and translation quality (BLEU). as well as the form of the generated subtitles (conformity to length/reading speed constraints and plausibility of segmentation), but also the consistency of the captions and subtitles in terms of number of subtitle blocks and lexical content in each subtitle. 

While
%\mn{Although}
some metrics for evaluating transcription-translation consistency have been proposed in \cite{sperber-etal-2020-consistent}, these do not capture the 
%particularities
peculiarities of subtitling. The goal for consistent captions/subtitles is having the same structure (same number of subtitle blocks) and same content (each pair of caption-subtitle blocks has the same lexical content).
%\mt{How is this approach similar to Sperber 2020? If not, you can stress that this is done for the first time.} 
%In order to obtain full lexical consistency, 
%they are both divided in two lines, containing the same lexical content...BLA BLA...THIS CAN BE THE PLACE FOR SHOWING WHAT  BLOCK AND LINES ACTUALLY ARE...ALSO MAKING THE NOTION OF CONSISTENCY EASY TO UNDERSTAND.
Consider the following example:

%Talk ID: 000003
\begin{figure}[h]
    \small
 0:00:50,820, 00:00:53,820\\
  To put the assumptions very clearly: \\
  Enonçons clairement nos hypothèses : le capitalisme, \\
  
  00:00:53,820, 00:00:57,820\\
  capitalism, after 150 years, has become acceptable,\\
  après 150 ans, est devenu acceptable, au même titre\\
  
  00:00:58,820, 00:01:00,820\\
  and so has democracy. \\
  que la démocratie. %Si nous prenons le monde tel qu'il était en 1945, et
 
    %\caption{Source captions (En) and Target subtitles (Fr) for the same subtitle template.}
    \label{fig:consub}
\end{figure}
In the example above, three blocks appear sequentially on the screen based on timestamp information, and each of them contains one line of text in English (caption) and French (subtitle).
Since the source utterance is split across the same number of blocks (3), the captions and subtitles have the same structure. However, the captions and subtitles do not have the same lexical content. The first block contains the French words
\textit{le capitalisme}, which appear in the second block for the English captions. Similarly, \textit{au même titre} corresponds to the third block in relation to the captions. This is problematic because terms do not appear in the same blocks (e.g. capitalism), and also leads to sub-optimal segmentation, since the French subtitles are not complete semantic units (logical completion occurs after \textit{hypothèses} and \textit{acceptable}). 

We hence define the consistency between captions and subtitles based on two aspects: the structural and the lexical consistency. 
Structural consistency refers to the way subtitles are distributed on a video. %, therefore the number of lines and blocks. 
% In order to be structurally consistent, each source utterance should have the same number of subtitle blocks between the captions and the subtitles. 
%\mn{
In order to be structurally consistent, captions and subtitles for each source utterance should be split across the same number of blocks. %, possibly divided in the same number of lines.} 
This is a prerequisite for bilingual subtitles, since  each caption-subtitle pair has the same timestamps. In other words, the captions and subtitles should appear and disappear simultaneously. Therefore, we define \textbf{structural consistency} as the percentage of utterances having the same number of blocks between captions and subtitles.%:
%\begin{equation}
%\small
%    struc = \frac{\#utterances(<eob>_{capt} = <eob>_{sub})}{\#tot\_utterances}
%\end{equation}

The second aspect of subtitling consistency is lexical consistency.  Lexical consistency means that each caption-subtitle pair has the same lexical content. It is particularly important for ensuring synchrony between the content displayed in the captions and subtitles. This facilitates language learning, when terms appear in similar positions, and quality assurance, as it is easier to spot errors in parallel text. 
We define \textbf{lexical consistency} as the percentage of words in each caption-subtitle pair that are aligned to words belonging in the same block. In our example, there are six tokens of the subtitles which are not aligned to captions of the same block: \textit{le capitalisme , au même titre}. %\mt{
For obtaining this score, we compute the number of words in each caption aligned to the corresponding subtitle and vice versa. For each caption-subtitle pair, this process results in two lexical consistency scores: Lex$_{caption \rightarrow subtitle}$ and Lex$_{subtitle \rightarrow  caption}$,\ where, in the former, the number of aligned words is normalised by the number of words in the caption, while, in the latter, by the number of words in the subtitle. These two quantities are then averaged into a single value (Lex$_{pair}$). The corpus-level lexical consistency is obtained by averaging the Lex$_{pair}$ of all caption-subtitle pairs in the test set.

%\begin{equation}
%    lex_{cap \rightarrow sub} = \frac{\sum_{b=1}^{B} \#( w_{capt}^{b} \textrm{ aligned\_with } w_{sub}^{b})} {\# w_{capt}}
%\end{equation}}
%\mt{where $B$ is total number of blocks in the subtitle. [ALINA: if we compute it from cap to sub, the number of blocks is in the subtitle ]
%  We average the aligned words over translation directions and normalise by the number of words of the longest sequence between caption/subtitle. 
% \begin{equation}
%     lex_{i} = \frac{\sum_{b=1}^{B} \#( w_{capt}^{b} \textrm{ aligned\_with } w_{sub}^{b})} {max(w_{capt}, w_{sub})}
% \end{equation}
% where $B$ is total number of blocks in $i$. 
% Counts of both metrics are aggregated and normalised at corpus level. 

\section{Experimental setting} \label{sec:expset}

\subsection{Data}
For our experiments we use MuST-Cinema \cite{karakanta-etal-2020-must}, an ST corpus compiled from subtitles of TED talks. For a sound comparison with \citet{karakanta-etal-2020-42}, we conduct the experiments on 2 language pairs, English$\rightarrow$French and English$\rightarrow$German. The breaks between subtitles are marked with special symbols, $<$eob$>$ for breaks between blocks of subtitles and $<$eol$>$ for new lines inside the same block. The training data contain 408 and 492 hours of pre-segmented audio (229K and 275K sentences) for German and French respectively. For tuning and evaluation we use the official development and test sets. We expect the captions and subtitles of TED Talks to have high consistency, since the captions %are created first and 
serve as the basis for translating the speech in target subtitles.

The text data is segmented into sub-words with Sentencepiece \cite{kudo-richardson-2018-sentencepiece} with the unigram setting. In line with recent works in ST, we found that a small vocabulary size is beneficial for the performance of ST models. Therefore, we set a shared vocabulary of 1024 for all models except the MT component of the cascade, where vocabulary size is set to 24k. The special symbols $<$eob$>$ and $<$eol$>$ are kept as a single token. 

For the audio input, we use 40-dimensional log Mel filterbank speech features. The ASR encoder was pretrained on the IWSLT 2020 data, i.e. Europarl-ST \cite{europarlst},
Librispeech \cite{librispeech}, How2 \cite{sanabria18how2}, Mozilla Common-Voice,\footnote{\url{https://voice.mozilla.org/}} MuST-C \cite{mustc}, and the  ST TED corpus.\footnote{\url{http://iwslt.org/doku.php?id=offline\_speech\_translation}}

\subsection{Model training}
The ASR and ST models are trained using the same settings. The architecture used is S-Transformer, \cite{digangi-2019-stransformer}, an ST adaptation of Transformer, which has been shown to achieve high performance on different speech translation benchmarks. Following state-of-the-art systems~\cite{potapczyk-przybysz-2020-srpols,gaido-etal-2020-end}, we do not add 2D self-attentions.
%.
The size of the encoder is set to 11 layers, and to 4 layers for the decoder.
The ASR model used to pretrain the encoder, instead, has 8 encoder and 6 decoder layers. The additional 3 encoder layers are initialised randomly, similarly to the adaptation layer proposed by \citet{bahar2019comparative}.
As distance penalty, we choose the logarithmic distance penalty. We optimise using Adam~\cite{kingma2014adam} (betas 0.9, 0.98), 4000 warm-up steps with initial learning rate of 0.0003, and learning rate decay with the inverse square root of the iteration. We apply label smoothing of 0.1, and dropout~\cite{srivastava2014dropout} is set to 0.2. We further use SpecAugment \cite{Park_2019}, a technique for online data augmentation, with augment rate of 0.5. Training is completed when the validation perplexity does not improve for 3 consecutive epochs. %, where primary perplexity for the cascade and is the transcription perplexity and translation perplexity for the joint models.

The MT component is based on the Transformer architecture (big)~\cite{vaswani2017attention} with similar settings to the original paper. Since the ASR component outputs punctuation, no other pre-processing (except for BPE) is applied to the training data. In order to ensure a fair comparison with the direct and joint models, the MT component is trained only on MuST-Cinema data.

All experiments are run with the fairseq toolkit \cite{ott-etal-2019-fairseq}. Training is performed on two K80 GPUs with 11 GB memory and models converged in about five days. Our implementation of the \texttt{DirMu}, \texttt{Tri} and \texttt{2ST} models is publicly available at: \url{https://github.com/mgaido91/FBK-fairseq-ST/tree/acl_2021}
%is based on Pytorch and 
%will be made publicly available upon acceptance of the paper.

\subsection{Evaluation}
\label{sec:eval}
We evaluate three aspects of the automatically generated captions and subtitles: 1) quality, 2) form, and 3) consistency. 
For \textbf{quality} of transcription we compute WER on unpunctuated, lowercased output, while for quality of translation we use SacreBLEU \cite{post-2018-call}.\footnote{\texttt{BLEU+c.mixed+\#.1+s.exp+tok.13a+v.1.4.3}} We report scores computed at the level of utterances, where the output sentences contain subtitle breaks. A break symbol is considered as another token contributing to the score.

For evaluating the \textbf{form} of the subtitles, we focus on the conformity to the subtitling constraints of length and reading speed, as well as proper segmentation, as proposed in \cite{karakanta2019subtitling}. We compute the percentage of subtitles conforming to a maximum length of 42 characters/line and a maximum reading speed of 21 characters/second.\footnote{In line with the TED Talk guidelines: \url{https://www.ted.com/participate/translate/guidelines}} The plausibility of segmentation is evaluated based on syntactic properties. Subtitle breaks should be placed in such a way that keeps syntactic and semantic units together. %This means that each subtitle block and subtitle line should be a complete linguistic unit. 
For example, an adjective should not be separated from the noun it describes. We consider as plausible only those breaks following punctuation marks %(where a logical completion is reached) 
or those between a content word (chunk) and a function word (chink). We obtain Universal Dependencies\footnote{\url{https://universaldependencies.org/}} PoS-tags using the Stanza toolkit \cite{qi-etal-2020-stanza} and calculate the percentage of break symbols falling either in the punctuation or the content-function groups as plausible segmentation.

Lastly, we evaluate structural and lexical \textbf{consistency} between the generated captions and corresponding subtitles, as described in Section~\ref{sec:consistency-def}. %For structural consistency, we estimate the percentage of sentences having the same number of subtitle blocks, while for lexical consistency we compute the percentage of aligned words not belonging to the same subtitle block. 
Word alignments are obtained using fast$\_$align \cite{dyer-etal-2013-simple} on the concatenation of MuST-Cinema training data and the system outputs. Text is tokenised using Moses tokeniser and the consistency percentage is computed on tokenised text.

\section{Results}
\label{sec:results}

\begin{table*}[ht]
\small
    \centering
    \begin{tabular}{l|rr|rrr|rr}\toprule
        En$\rightarrow$Fr	& WER	& SacreBLEU  & Length & Read\_speed  & Segment.  & Struc.  & Lex. \\ \midrule
         Cas & 19.69 & \textbf{26.9} & .94 / .93	& .85 / .70  & .86 / .82 &  \textbf{.98} &	\textbf{.99}\\
         DirInd & 19.69 & 24.0 & .94 / \textbf{.94}	& .85 / \textbf{.73} & .86 / \textbf{.84} & .75	& .86 \\
         DirMu & \textbf{17.73} & 25.2 & \textbf{.95} / .93	& .85 / \textbf{.73} & \textbf{.87} / .80 & .77 &	.87 \\
         2ST & 19.05 & 25.6 & \textbf{.95} / \textbf{.94} & .85 / .71 & \textbf{.87} / .82 & .83 & .84\\
         Tri & 18.93  & 25.3 & .93 / .91	& .85 / .72 & \textbf{.87} / .82 & .82	& .92\\  
         %\midrule
         %Ref & & & 100\%/100\% & 85\% / 80\% & 83\% / 75\% & 67\% & 92\% & 11\%\\ 
         \bottomrule
         En$\rightarrow$De	&   \\ \midrule
         Cas & 18.52	& \textbf{19.9} & .94 / .90	& .62 / .58 & .86 / .76 & \textbf{.95}	& \textbf{.96} \\
         DirInd & 18.52 & 18.1 & .94 / \textbf{.92}	& .62 / .59 & .86 / \textbf{.78} & .73	& .86 \\
         DirMu & \textbf{16.95}	& 18.7 & \textbf{.95} / \textbf{.92}	& .62 / .59 & \textbf{.87} / .73 & .75	& .82 \\
         2ST & 18.93  & \textit{19.6} & .94 / \textbf{.92} & .62 / \textbf{.60} & .86 / .76 & .82 & .81 \\
         Tri & 19.10 & \textit{19.8} & .93 / \textbf{.92}	& .62 / \textbf{.60} & \textbf{.87} / .76 & .78	& .91 \\  %\midrule
         %Ref & & & 100\%/100\% & 74\% / 72\% & 83\% / 66\% & 66\% & 92\% & 4\% \\  
         \bottomrule
    \end{tabular}
    \caption{Results for quality (WER and BLEU), subtitling conformity (Length, Reading speed and Segmentation), and subtitling consistency (Structural and Lexical) for model outputs for French and German. Conformity scores are reported for captions / subtitles. \textbf{Bold} denotes the best score. Results that are not statistically significant -- according to pairwise bootstrap resampling \cite{koehn-2004-statistical-sign}, p$<$0.05 -- than the best score are reported in \textit{italics}.}
    \label{tab:results}
\end{table*}

\subsection{Transcription/Translation quality}

We first examine the quality of the systems' outputs. The first two columns of Table~\ref{tab:results} show the WER and SacreBLEU score for the examined models. 

In terms of \textbf{transcription quality}, \texttt{DirMu} (Multitask Direct -- see Section \ref{subsec:models}) obtains the lowest WER for both languages (17.73 for French and 16.95 for German). As far as the rest of the models are concerned, there is a different tendency for French and German. \texttt{Tri} (Triangle) and \texttt{2ST} (Two-Stage) perform equally better than the \texttt{Cas}/\texttt{DirInd} for French, while the \texttt{Cas}/\texttt{DirInd} have higher transcription quality than \texttt{Tri} and \texttt{2ST} for German. An explanation for this incongruity is that these two models perform coupled inference, therefore the benefit of the joint decoding for the transcription can be related to similarities in terms of vocabulary between the two languages. Since French has a higher vocabulary similarity to English, with many words in TED Talks being cognates (e.g. specialised terminology), it is possible that joint decoding favours the transcription for French but not for German. %infor from translation may not be useful to improve the transcription, while 2 tasks can enhance the encoder (DirMu).

When it comes to \textbf{translation quality}, \texttt{Cas} outperforms all other models for French with 26.9 BLEU points, while the differences are not statistically significant among \texttt{DirMu}, \texttt{2ST} and \texttt{Tri}. For German, however, 
\texttt{Cas}, \texttt{2ST} and \texttt{Tri} perform on par. The model obtaining the lowest scores is 
\texttt{DirInd}.
This finding confirms our hypothesis that joint decoding, despite being more complex, improves translation quality thanks to the knowledge shared between the two tasks at decoding time.

In comparison to previous works, our transcription results are contrary to \citet{sperber-etal-2020-consistent}, who obtained the lowest WER with the 
cascade and direct independent models.
However, for translation quality our best models are 
\texttt{Cas}, \texttt{2ST} and \texttt{Tri},
as in previous work. Moreover, in line with \citet{anastasopoulos-chiang-2018-tied}, the gains for \texttt{Tri} are higher for translation than for transcription. 
Comparing the BLEU score of our \texttt{DirInd} models to the models in \cite{karakanta-etal-2020-42}, we found that our models achieve higher performance with 20.07 BLEU compared to 18.76 for French and 13.55 compared to 11.92 for German. %This difference could be due to the smaller vocabulary size of 1k instead of 8k. 

All in all, we found that coupled-inference, supported by \texttt{Cas}, \texttt{2ST} and \texttt{Tri}, improves translation but not transcription quality. On the contrary, multi-tasking as in \texttt{DirMu} is beneficial for transcription, possibly because of a reinforcement of the speech encoder. However, could this improvement come at the expense of conformity to the subtitling constraints? 

\subsection{Subtitling conformity}\label{subs:res-confo}
%In the previous section, we confirmed that joint decoding benefits transcription and translation quality.
Columns 3-5 of Table~\ref{tab:results} show the percentage of captions/subtitles conforming to the length, reading speed and segmentation constraints discussed in Section~\ref{sec:eval}.
We observe that joint decoding does not lead to significant losses in conformity. Specifically, the captions generated by \texttt{DirMu} have the highest conformity in terms of length (95\%), reading speed (85\% and 62\%) and segmentation quality (87\%). Moreover, the high conformity score for \texttt{DirMu} correlates with the low WER, showing that quality goes hand-in-hand with conformity.

For the conformity of the target language subtitles, instead, the picture is different. Even though the differences are not large, \texttt{Cas} has lower conformity to length (93\% and 90\%) and reading speed (70\% and 58\%). The segmentation scores show that, despite their high translation quality, the systems featuring coupled inference (\texttt{Cas}, \texttt{Tri} and \texttt{2ST}) are constrained by the structure of the captions and segment subtitles in positions which are not optimal for the target language norms %, as shown by the scores for the segmentation 
(82\% and 76\%)%, leading, in turn, to higher length and faster reading speed
. \texttt{DirInd}, on the contrary, has higher conformity compared to the other models (94\% and 92\% for length, 73\% and 59\% for reading speed), as well as segmentation quality (84\% and 78\%). \texttt{DirInd}  is left to determine the most plausible segmentation for the target language without being bound by consistency constraints from the source. The lowest segmentation quality of subtitles is achieved by \texttt{DirMu} (80\% and 73\%). %Extreme sharing of all weights in \texttt{DirMu} hurts the segmentation of subtitles, which is determined by target language syntactic norms. 

We can conclude that the quality improvements of coupled inference and multi-tasking come with a slight compromise of subtitling conformity, as a result of loss of flexibility in decoding. %the conformity metrics paint a spectrum between the high flexibility of independent decoding and the extreme sharing of \texttt{DirMu}.

\subsection{Subtitling consistency}\label{subs:res-cons}
The last two columns of Table~\ref{tab:results} present the results for the subtitling consistency.

%\paragraph{Structural consistency:}
In terms of \textbf{Structural consistency} (Struc.), the model achieving the highest scores  is \texttt{Cas}, with 98\% and 95\% of the utterances being distributed along the same number of blocks. As expected, the lowest structural  consistency is achieved by \texttt{DirInd} (75\% and 73\%), which determines independently the positions of the block symbols. Among the joint models, \texttt{Tri} outputs captions and subtitles with higher consistency than \texttt{DirMu}, but both are outperformed by \texttt{2ST} (83\% and 82\%). Our hypothesis is that by attending only the caption decoder, \texttt{2ST} behaves similarly to the cascade, and the translation decoder better replicates the block structure. We noted that the reference captions and subtitles have lower consistency (92\% both for French and German) than the cascade. This shows that the cascade copies the same $<$eob$>$ tokens and achieves extreme structural consistency, which is a desideratum for our study case but may be harmful 
%not be desired as important 
in other scenarios, since it leads to lower conformity %to the target language norms 
(see Section~\ref{subs:res-confo}). Indeed, in scenarios where consistency is not a key, subtitlers should have the flexibility to adjust subtitling segmentation %time codes as well as delete, merge and split the subtitles
to suit the needs of their target languages \cite{oziemblewska-2020-templ}.%\mt{Can the score of the reference depend on the fact that the translations can be free so they do not match the structure of the source? I was wondering if adding this information here can hurt other parts of the paper? }

%\paragraph{Lexical consistency:}
The \textbf{Lexical consistency} (Lex.) results show that \texttt{Cas} is the model with the highest content overlap in parallel caption-subtitle blocks with 99\% and 96\% of the words being aligned to the same block. As with the structural consistency, the lexical consistency of the cascade is higher than the references
%, which show a lexical consistency of 
(95\% for French, 86\% for German). The direct model with the highest lexical consistency is \texttt{Tri} (92\% and 91\%). Interestingly, despite its high structural consistency, \texttt{2ST} does not distribute the content consistently in the parallel blocks, achieving the lowest conformity (81\%). The \texttt{DirMu} also achieves lower consistency than \texttt{DirInd} for German (82\% compared to 86\%) but not for French (87\% compared to 86\%). It is worth noting that lexical consistency is generally lower for German than for French. Indeed, a 100\% lexical consistency between subtitles in languages with different word order is not always feasible or even appropriate. For example, the main verb in an English subordinate clause appears in the second position while in German at the end of the sentence. In order to adhere to grammatical rules, words in subtitles of different languages often have inter-block reordering. Therefore, the balance between flexibility and consistency is manifested here as a compromise between grammaticallity and preservation of the same lexical content on each pair of subtitles.

To sum up, the results of structural consistency show that the models are able to preserve the block structure between captions and subtitles in more than 75\% of the utterances. In addition, the high lexical consistency shows that the block symbols are not inserted randomly, but placed in a way that preserves the same lexical content in the parallel blocks.%, as shown by the high lexical consistency. 

All in all, our results show that the evaluation of captions and subtitles is a multifaceted process that needs to be addressed from multiple aspects: quality, conformity and consistency. Missing one of the three can lead to wrong conclusions. For instance, only considering quality and consistency could lead to disregard the importance of conformity and consider independent solutions an obsolete technology. 
Secondly, among the Direct architectures, the use of techniques that allow linking the generation process of captions and subtitles helps to achieve overall better quality and consistency than independent decoding, with a slight discount in conformity, especially for the target subtitles. Between the \texttt{DirMu}, \texttt{2ST} and \texttt{Tri}, there  is not a model that outperforms all the others in all the metrics, so the choice mainly depends on the application scenario. 
Lastly, comparing the Cascade and the Direct, the Cascade seems to be the best choice, but recent advancements in Direct approaches result in competitive solutions with increased efficiency of maintaining one model for both tasks.
%Flexibility, properties and needs of each language. \mn{\textbf{???? to be completed}}

\section{Analysis} \label{sec:analysis}

\subsection{Evaluation of Lexical Consistency}
In this section, we test the reliability of the lexical consistency metric. The metric depends on the successful word alignment, which, especially for low quality text, might be sub-optimal. We therefore manually count the number of words in the subtitles which do not appear in the corresponding captions. The task is performed on the first 347 sentences of the output of \texttt{DirMu} for French and German. We then estimate the mean absolute error between the consistency metric computed using the manual and the automatic alignments.
%\st{the manual annotation and the lexical consistency metric.} \mt{the MAE should be computed on the difference of the same metric, computed in different conditions As it is written now, you compute the MAE between the manual annotation and the score produced by the metric.} 
As an additional step, we compute how often the automatic and the manual annotations agree in their judgement of consistent/non-consistent content in each block. 

The mean absolute error between the manually and the automatically computed score is .08 for French and .11 for German. The metric may not be able to account for very small score differences between systems, however, when inspecting the differences between manual and automatic annotation we noticed that most errors appear in very low quality outputs or where lexical content was missing, and lead to a misalignment of only a few words. These cases were in fact challenging even for the human annotator. Instead, the agreement in the consistent/not-consistent judgement is high, with .85 for French and .75 for German. Considering the difficulty of aligning sentences belonging to languages with different word ordering, and the lower quality of German outputs, it is not surprising that the word aligner from English to German affects more our metric%when dealing with these two languages
. However, these results show that the real impact is moderate and the metric is consistent with the human judgements in the majority of cases. %Indeed, more investigations are needed to better understand the potential of the Lexical consistency evaluation metric.
%Fr 0.08 agreement 0.85
%De 0.11 agreement 0.75

\subsection{Does structural consistency extend to line breaks?}
But what happens with the line breaks? Does a one-line %\mt{liner or line? } 
caption correspond to a one-line subtitle in the output of our models? Having the same number of lines between caption and subtitle blocks is a more challenging scenario, since the subtitles tend to expand because of different length ratios between languages and translation strategies such as explicitation. For instance, for the target languages considered in this work (French and German) the length of the target subtitles when subtitling from English has been reported to be 5\%-35\% higher.%\cite{kwintessential,aranchodoc,andiamo}
\footnote{\url{https://www.andiamo.co.uk/resources/expansion-and-contraction-factors/} \url{http://www.aranchodoc.com/wp-content/uploads/2017/07/Text-Expansion-Contraction-Chart3.png} \url{https://www.kwintessential.co.uk/resources/expansion-retraction}} If structural consistency is enforced to line breaks, it may compromise either the quality of the translation or the conformity to the subtitling constraints. In case of a one-liner caption, important information may be not rendered in the corresponding subtitle in order to match a shorter length of the caption, or the length constraint will be violated since the longer subtitle will not be adequately segmented in two lines. %However, we posit that the same number of lines inside a block may highlight the parallel structure and facilitate reading. 
In order to ensure that our models do not push the structural consistency to an extreme, we compute the percentage of caption-subtitle blocks having the same number of lines.

\begin{table}[ht]
    \centering \small
    \begin{tabular}{c|ccccc|c}\toprule
        & Cas & DirInd & DirMu & 2ST & Tri & Ref \\ \midrule
    Fr & 67\% & 49\% & 54\% & 59\% & 57\% & 67\% \\
    De & 66\% & 47\% & 55\% & 53\% & 51\% & 66\% \\ \bottomrule
%        Model & Fr & De \\ \midrule
%Cas & 67\% & 66\% \\
%DirInd & 49\% & 47\% \\
%DirMu & 54\% & 55\% \\
%2ST & 59\% & 53\% \\
%Tri & 57\% & 51\% \\ \midrule
%Ref & 67\% & 66\% \\ \bottomrule
    \end{tabular}
    \caption{Percentage of subtitle blocks containing the same number of lines for  French and German outputs.}
    \label{tab:lines}
\end{table}

Table~\ref{tab:lines} confirms that caption and subtitle blocks do not always have the same number of lines, since only 67\% and 66\% of blocks in the caption/subtitle references have the same number of lines. When it comes to the models, the cascade exactly matches the percentages of the references, while the direct models have even lower percentage of equal number of lines. Among the direct models, again the \texttt{DirInd} shows the lowest similarity.
%This is an element that is probably learned by the training data. 
We observed that more line breaks were present in the target subtitles, which ensures length conformity, since the target subtitles expand (source-target character ratio of 0.91 for French and 0.93 for German). %Ref char-ratio: 96 DE, 99 Fr
Therefore, the fact that structural consistency allows for flexibility in relation to the number and position of line breaks is key to achieving high quality and conformity.

%\subsection{Analysis of Segmentation}
%\citet{karakanta-etal-2020-42} claimed that the two types of subtitle breaks (block break $<$eob$>$ versus line break $<$eol$>$) have different functions. While line breaks function is to split a long subtitle into two smaller pieces to fit the screen and therefore its position is determined by  two  factors:  achieving  a  more  or  less  equallength of the upper and the lower subtitle line andinserting the break in a position such that syntacticunits are kept together. Subtitle blocks on the other hand correspond to larger pauses and therefore their position was more determined by the natural rhythm of the speech. Here, contrary to the accumulated Segmentation results in Table~\ref{tab:results}, we explore whether there are differences among the models in their accuracy of segmentation, distinguishing between the two break types.

\section{Conclusions} \label{sec:conclusion}
In this work we explored joint generation of captions and subtitles as a way to increase efficiency and consistency in scenarios where this property is a desideratum. To this aim, we proposed metrics for evaluating subtitling consistency, tailored to the structural peculiarities of this type of translation. We found that coupled inference, either by models supporting end-to-end training (\texttt{2ST}, \texttt{Tri}) or not (\texttt{Cas}), leads to quality and consistency improvements, but with a slight degradation of the conformity to target subtitle constraints. The final architectural choice depends on the flexibility versus conformity requirements of the application scenario.

The findings of this work have provided initial insights related to the joint generation of captions and subtitles. One future research direction is towards improving the quality of generation by using more recent, higher-performing ST architectures. For example, \citet{lui-et-al-2020-synchronous} extended the notion of the dual decoder by adding an interactive attention mechanism which allows the two decoders to exchange information and learn from each other, while synchronously generating transcription and translation. \citet{le-etal-2020-dual} proposed two variants of the dual decoder of \citet{lui-et-al-2020-synchronous}, the cross and parallel dual decoder, and experimented with multilingual ST. While neither of these works reported results on consistency, we expect that they are relevant to our scenario and have the potential of jointly generating multiple language/accessibility versions with high consistency.
Moving beyond generic architectures, in the future we are planning to experiment with tailored architectures for improving consistency between automatically generated captions and subtitles. One important insight emerging from this work is that different degrees of conformity are required, or even appropriate, depending on the application scenario and languages involved. Given these challenges, we are aiming at developing approaches which allow for tuning the output to the desired degree of conformity, whether lexical, structural or both.
We hope that this work will contribute to the line of research efforts towards improving efficiency and quality of automatically generated captions and subtitles.
%\section*{Acknowledgments}

\bibliographystyle{acl_natbib}
\bibliography{anthology,acl2021}

\begin{thebibliography}{47}
\expandafter\ifx\csname natexlab\endcsname\relax\def\natexlab#1{#1}\fi

\bibitem[{Anastasopoulos and Chiang(2018)}]{anastasopoulos-chiang-2018-tied}
Antonios Anastasopoulos and David Chiang. 2018.
\newblock \href {https://doi.org/10.18653/v1/N18-1008} {Tied multitask learning
  for neural speech translation}.
\newblock In \emph{Proceedings of the 2018 Conference of the North {A}merican
  Chapter of the Association for Computational Linguistics: Human Language
  Technologies, Volume 1 (Long Papers)}, pages 82--91, New Orleans, Louisiana.
  Association for Computational Linguistics.

\bibitem[{Armstrong et~al.(2006)Armstrong, Caffrey, and
  Flanagan}]{armstrong-2006-dvd}
Stephen Armstrong, Colm Caffrey, and Marian Flanagan. 2006.
\newblock \href {http://www.mt-archive.info/MuTra-2006-Armstrong-1.pdf}
  {{Translating DVD Subtitles from English-German and English-Japanese Using
  Example-Based Machine Translation}}.
\newblock In \emph{MuTra 2006—Audiovisual Translation Scenarios: Conference
  Proceedings}.

\bibitem[{Bahar et~al.(2019)Bahar, Bieschke, and Ney}]{bahar2019comparative}
Parnia Bahar, Tobias Bieschke, and Hermann Ney. 2019.
\newblock {A Comparative Study on End-to-end Speech to Text Translation}.
\newblock In \emph{Proceedings of International Workshop on Automatic Speech
  Recognition and Understanding (ASRU)}, pages 792--799, Sentosa, Singapore.

\bibitem[{Bérard et~al.(2016)Bérard, Pietquin, Servan, and
  Besacier}]{berard_2016}
Alexandre Bérard, Olivier Pietquin, Christophe Servan, and Laurent Besacier.
  2016.
\newblock {Listen and Translate: A Proof of Concept for End-to-End
  Speech-to-Text Translation}.
\newblock In \emph{NIPS Workshop on end-to-end learning for speech and audio
  processing}, Barcelona, Spain.

\bibitem[{Cattoni et~al.(2020)Cattoni, Di~Gangi, Bentivogli, Negri, and
  Turchi}]{mustc}
Roldano Cattoni, Mattia~A. Di~Gangi, Luisa Bentivogli, Matteo Negri, and Marco
  Turchi. 2020.
\newblock {MuST-C: A Multilingual Corpus for end-to-end Speech Translation}.
\newblock Computer Speech \& Language Journal.
\newblock Doi: https://doi.org/10.1016/j.csl.2020.101155.

\bibitem[{Cintas and Remael(2007)}]{Avt2007sub}
Jorge~Diaz Cintas and Aline Remael. 2007.
\newblock \emph{Audiovisual Translation: Subtitling}.
\newblock Translation practices explained. Routledge.

\bibitem[{Di~Gangi et~al.(2019)Di~Gangi, Negri, and
  Turchi}]{digangi-2019-stransformer}
Mattia~Antonino Di~Gangi, Matteo Negri, and Marco Turchi. 2019.
\newblock {Adapting Transformer to End-to-end Spoken Language Translation}.
\newblock In \emph{INTERSPEECH}, pages 1133--1137.

\bibitem[{Dyer et~al.(2013)Dyer, Chahuneau, and Smith}]{dyer-etal-2013-simple}
Chris Dyer, Victor Chahuneau, and Noah~A. Smith. 2013.
\newblock \href {https://www.aclweb.org/anthology/N13-1073} {A simple, fast,
  and effective reparameterization of {IBM} model 2}.
\newblock In \emph{Proceedings of the 2013 Conference of the North {A}merican
  Chapter of the Association for Computational Linguistics: Human Language
  Technologies}, pages 644--648, Atlanta, Georgia. Association for
  Computational Linguistics.

\bibitem[{Etchegoyhen et~al.(2014)Etchegoyhen, Bywood, Fishel, Georgakopoulou,
  Jiang, van Loenhout, del Pozo, Maučec, Turner, and
  Volk}]{Etchegoyhen-2014-sumat}
Thierry Etchegoyhen, Lindsay Bywood, Mark Fishel, Panayota Georgakopoulou, Jie
  Jiang, Gerard van Loenhout, Arantza del Pozo, Mirjam~Sepesy Maučec, Anja
  Turner, and Martin Volk. 2014.
\newblock \href
  {http://www.lrec-conf.org/proceedings/lrec2014/pdf/463_Paper.pdf} {{Machine
  Translation for Subtitling: A Large-Scale Evaluation}}.
\newblock In \emph{Proceedings of the 9th International Conference on Language
  Resources and Evaluation (LREC)}, pages 46--53.

\bibitem[{Gaido et~al.(2020)Gaido, Di~Gangi, Negri, and
  Turchi}]{gaido-etal-2020-end}
Marco Gaido, Mattia~A. Di~Gangi, Matteo Negri, and Marco Turchi. 2020.
\newblock \href {https://doi.org/10.18653/v1/2020.iwslt-1.8} {End-to-end
  speech-translation with knowledge distillation: {FBK}@{IWSLT}2020}.
\newblock In \emph{Proceedings of the 17th International Conference on Spoken
  Language Translation}, pages 80--88, Online. Association for Computational
  Linguistics.

\bibitem[{García(2017)}]{garcia-2017-bisubsf}
Boni García. 2017.
\newblock \href {https://doi.org/https://doi.org/10.1002/cae.21814} {Bilingual
  subtitles for second-language acquisition and application to engineering
  education as learning pills}.
\newblock \emph{Computer Applications in Engineering Education},
  25(3):468--479.

\bibitem[{Georgakopoulou(2019)}]{georgakopoulou-19}
Panayota Georgakopoulou. 2019.
\newblock {Template files: The Holy Grail of subtitling}.
\newblock \emph{Journal of Audiovisual Translation}, 2(2):137--160.

\bibitem[{Gottlieb(2004)}]{gottlieb-2004}
Henrik Gottlieb. 2004.
\newblock Subtitles and international anglification.
\newblock \emph{Nordic Journal of English Studies}, 3:219--230.

\bibitem[{Guichon and McLornan(2008)}]{guichon-2008-learn}
Nicolas Guichon and Sinead McLornan. 2008.
\newblock {The effects of multimodality on l2 learners: Implications for call
  resource design}.
\newblock \emph{System}, pages 85--93.

\bibitem[{Iranzo-S\'{a}nchez et~al.(2020)Iranzo-S\'{a}nchez,
  Silvestre-Cerd\`{a}, Jorge, Rosell\'{o}, Adri\`{a}, Sanchis, Civera, and
  Juan}]{europarlst}
Javier Iranzo-S\'{a}nchez, Joan~Albert Silvestre-Cerd\`{a}, Javier Jorge,
  Nahuel Rosell\'{o}, Gim\'{e}nez. Adri\`{a}, Albert Sanchis, Jorge Civera, and
  Alfons Juan. 2020.
\newblock \href {https://ieeexplore.ieee.org/document/9054626} {{Europarl-ST: A
  Multilingual Corpus For Speech Translation Of Parliamentary Debates}}.
\newblock In \emph{Proceedings of 2020 IEEE International Conference on
  Acoustics, Speech and Signal Processing (ICASSP)}, pages 8229--8233,
  Barcelona, Spain.

\bibitem[{Kano et~al.(2017)Kano, Sakti, and Nakamura}]{kano-2017-2stage}
Takatomo Kano, Sakriani Sakti, and Satoshi Nakamura. 2017.
\newblock \href
  {https://doi.org/I:https://doi.org/10.21437/Interspeech.2017-944}
  {{Structured-based Curriculum Learning for End-to-end English-Japanese Speech
  Translation}}.
\newblock In \emph{Annual Conference of the International Speech Communication
  Association (InterSpeech)}, pages 2630--2634.

\bibitem[{Karakanta et~al.(2019)Karakanta, Negri, and
  Turchi}]{karakanta2019subtitling}
Alina Karakanta, Matteo Negri, and Marco Turchi. 2019.
\newblock \href {http://ceur-ws.org/Vol-2481/paper38.pdf} {{Are Subtitling
  Corpora really Subtitle-like?}}
\newblock In \emph{Sixth Italian Conference on Computational Linguistics,
  CLiC-It}.

\bibitem[{Karakanta et~al.(2020{\natexlab{a}})Karakanta, Negri, and
  Turchi}]{karakanta-etal-2020-42}
Alina Karakanta, Matteo Negri, and Marco Turchi. 2020{\natexlab{a}}.
\newblock \href {https://doi.org/10.18653/v1/2020.iwslt-1.26} {Is 42 the answer
  to everything in subtitling-oriented speech translation?}
\newblock In \emph{Proceedings of the 17th International Conference on Spoken
  Language Translation}, pages 209--219, Online. Association for Computational
  Linguistics.

\bibitem[{Karakanta et~al.(2020{\natexlab{b}})Karakanta, Negri, and
  Turchi}]{karakanta-etal-2020-must}
Alina Karakanta, Matteo Negri, and Marco Turchi. 2020{\natexlab{b}}.
\newblock \href {https://www.aclweb.org/anthology/2020.lrec-1.460}
  {{M}u{ST}-cinema: a speech-to-subtitles corpus}.
\newblock In \emph{Proceedings of the 12th Language Resources and Evaluation
  Conference}, pages 3727--3734, Marseille, France. European Language Resources
  Association.

\bibitem[{Kingma and Ba(2015)}]{kingma2014adam}
Diederik Kingma and Jimmy Ba. 2015.
\newblock Adam: A method for stochastic optimization.
\newblock In \emph{3rd International Conference for Learning Representations}.

\bibitem[{Koehn(2004)}]{koehn-2004-statistical-sign}
Philipp Koehn. 2004.
\newblock \href {https://www.aclweb.org/anthology/W04-3250} {{Statistical
  Significance Tests for Machine Translation Evaluation}}.
\newblock In \emph{Proceedings of the 2004 Conference on Empirical Methods in
  Natural Language Processing}, pages 388--395, Barcelona, Spain. Association
  for Computational Linguistics.

\bibitem[{Koponen et~al.(2020)Koponen, Sulubacak, Vitikainen, and
  Tiedemann}]{koponen-2020}
Maarit Koponen, Umut Sulubacak, Kaisa Vitikainen, and Jörg Tiedemann. 2020.
\newblock {MT for subtitling: User} evaluation of post-editing productivity.
\newblock In \emph{Proceedings of the 22nd Annual Conference of the European
  Association for Machine Translation (EAMT 2020)}, pages 115--124.

\bibitem[{Kudo and Richardson(2018)}]{kudo-richardson-2018-sentencepiece}
Taku Kudo and John Richardson. 2018.
\newblock \href {https://doi.org/10.18653/v1/D18-2012} {{S}entence{P}iece: A
  simple and language independent subword tokenizer and detokenizer for neural
  text processing}.
\newblock In \emph{Proceedings of the 2018 Conference on Empirical Methods in
  Natural Language Processing: System Demonstrations}, pages 66--71, Brussels,
  Belgium. Association for Computational Linguistics.

\bibitem[{Le et~al.(2020)Le, Pino, Wang, Gu, Schwab, and
  Besacier}]{le-etal-2020-dual}
Hang Le, Juan Pino, Changhan Wang, Jiatao Gu, Didier Schwab, and Laurent
  Besacier. 2020.
\newblock \href {https://www.aclweb.org/anthology/2020.coling-main.314}
  {Dual-decoder transformer for joint automatic speech recognition and
  multilingual speech translation}.
\newblock In \emph{Proceedings of the 28th International Conference on
  Computational Linguistics}, pages 3520--3533, Barcelona, Spain (Online).
  International Committee on Computational Linguistics.

\bibitem[{Liao et~al.(2020)Liao, Kruger, and Doherty}]{Liao2020TheIO}
Sixin Liao, Jan-Louis Kruger, and Stephen Doherty. 2020.
\newblock \href {https://www.jostrans.org/issue33/art_liao.pdf} {The impact of
  monolingual and bilingual subtitles on visual attention, cognitive load, and
  comprehension}.
\newblock \emph{The Journal of Specialised Translation}.

\bibitem[{Liu et~al.(2020)Liu, Zhang, Xiong, Zhou, He, Wu, Wang, and
  Zong}]{lui-et-al-2020-synchronous}
Yuchen Liu, Jiajun Zhang, Hao Xiong, Long Zhou, Zhongjun He, Hua Wu, Haifeng
  Wang, and Chengqing Zong. 2020.
\newblock \href
  {http://nlpr-web.ia.ac.cn/cip/ZongPublications/2020/AAAI2020_liuyuchen.pdf}
  {{Synchronous Speech Recognition and Speech-to-Text Translation with
  Interactive Decoding}}.
\newblock In \emph{The Thirty-Fourth AAAI Conference on Artificial Intelligence
  (AAAI-20)}. Association for the Advancement of Artificial Intelligence
  (www.aaai.org).

\bibitem[{Matusov et~al.(2019)Matusov, Wilken, and
  Georgakopoulou}]{matusov-etal-2019-customizing}
Evgeny Matusov, Patrick Wilken, and Yota Georgakopoulou. 2019.
\newblock \href {https://doi.org/10.18653/v1/W19-5209} {Customizing neural
  machine translation for subtitling}.
\newblock In \emph{Proceedings of the Fourth Conference on Machine Translation
  (Volume 1: Research Papers)}, pages 82--93, Florence, Italy. Association for
  Computational Linguistics.

\bibitem[{Melero et~al.(2006)Melero, Oliver, and Badia}]{melero-2006-etitle}
Maite Melero, Antoni Oliver, and Toni Badia. 2006.
\newblock \href
  {http://citeseerx.ist.psu.edu/viewdoc/download?doi=10.1.1.107.6011&rep=rep1}
  {{Automatic Multilingual Subtitling in the eTITLE Project}}.
\newblock In \emph{Proceedings of ASLIB Translating and the Computer 28}.

\bibitem[{Netflix(2021)}]{netflix-template}
Netflix. 2021.
\newblock Subtitle template timed text style guide.
\newblock
  https://partnerhelp.netflixstudios.com/hc/en-us/articles/219375728-English-Template-Timed-Text-Style-Guide.
\newblock Last accessed: 02/05/2021.

\bibitem[{Nikoli{\'{c}}(2015)}]{Nikolic2015}
Kristijan Nikoli{\'{c}}. 2015.
\newblock \href {https://doi.org/10.1057/9781137552891_11} {The pros and cons
  of using templates in subtitling}.
\newblock In Roc{\'i}o~Ba{\~{n}}os Pi{\~{n}}ero and Jorge~D{\'i}az Cintas,
  editors, \emph{Audiovisual Translation in a Global Context: Mapping an
  Ever-changing Landscape}, pages 192--202. Palgrave Macmillan UK, London.

\bibitem[{Nyberg and Mitamura(1997)}]{nyberg-1997-captions}
Eric Nyberg and Teruko Mitamura. 1997.
\newblock \href {http://www.lti.cs.cmu.edu/Research/Kant/PDF/mts6.pdf} {{A
  Real-Time MT System for Translating Broadcast Captions}}.
\newblock In \emph{Proceedings of the Sixth Machine Translation Summit}, pages
  51--57.

\bibitem[{Ott et~al.(2019)Ott, Edunov, Baevski, Fan, Gross, Ng, Grangier, and
  Auli}]{ott-etal-2019-fairseq}
Myle Ott, Sergey Edunov, Alexei Baevski, Angela Fan, Sam Gross, Nathan Ng,
  David Grangier, and Michael Auli. 2019.
\newblock \href {https://doi.org/10.18653/v1/N19-4009} {fairseq: A fast,
  extensible toolkit for sequence modeling}.
\newblock In \emph{Proceedings of the 2019 Conference of the North {A}merican
  Chapter of the Association for Computational Linguistics (Demonstrations)},
  pages 48--53, Minneapolis, Minnesota. Association for Computational
  Linguistics.

\bibitem[{Oziemblewska and Szarkowska(2020)}]{oziemblewska-2020-templ}
Magdalena Oziemblewska and Agnieszka Szarkowska. 2020.
\newblock \href {https://doi.org/10.1080/0907676X.2020.1791919} {{The quality
  of templates in subtitling. A survey on current market practices and changing
  subtitler competences}}.
\newblock \emph{Perspectives}, 0(0):1--22.

\bibitem[{Panayotov et~al.(2015)Panayotov, Chen, Povey, and
  Khudanpur}]{librispeech}
Vassil Panayotov, Guoguo Chen, Daniel Povey, and Sanjeev Khudanpur. 2015.
\newblock {Librispeech: An ASR corpus based on public domain audio books}.
\newblock In \emph{2015 IEEE International Conference on Acoustics, Speech and
  Signal Processing (ICASSP)}, pages 5206--5210, South Brisbane, Queensland,
  Australia.

\bibitem[{Park et~al.(2019)Park, Chan, Zhang, Chiu, Zoph, Cubuk, and
  Le}]{Park_2019}
Daniel~S. Park, William Chan, Yu~Zhang, Chung-Cheng Chiu, Barret Zoph, Ekin~D.
  Cubuk, and Quoc~V. Le. 2019.
\newblock \href {https://doi.org/10.21437/interspeech.2019-2680} {{SpecAugment:
  A Simple Data Augmentation Method for Automatic Speech Recognition}}.
\newblock \emph{Interspeech 2019}.

\bibitem[{Piperidis et~al.(2005)Piperidis, Demiros, and
  Prokopidis}]{piperidis-2005-musa}
Stelios Piperidis, Iason Demiros, and Prokopis Prokopidis. 2005.
\newblock \href
  {http://sifnos.ilsp.gr/musa/publications/Infrastructure\%20for\%20a\%20multilingual\%20subtitle\%20generation\%20system-final.pdf}
  {{Infrastructure for a Multilingual Subtitle Generation System}}.
\newblock In \emph{9th International Symposium on Social Communication}, pages
  24--28.

\bibitem[{Popowich et~al.(2000)Popowich, McFetridge, Turcato, and
  Toole}]{Popowich-2000-mtcaptiona}
Fred Popowich, Paul McFetridge, Davide Turcato, and Janine Toole. 2000.
\newblock \href
  {http://www.jstor.org/stable/20060451?seq=1#page_scan_tab_contents} {{Machine
  Translation of Closed Captions}}.
\newblock \emph{Machine Translation}, pages 311--341.

\bibitem[{Post(2018)}]{post-2018-call}
Matt Post. 2018.
\newblock \href {https://doi.org/10.18653/v1/W18-6319} {A call for clarity in
  reporting {BLEU} scores}.
\newblock In \emph{Proceedings of the Third Conference on Machine Translation:
  Research Papers}, pages 186--191, Brussels, Belgium. Association for
  Computational Linguistics.

\bibitem[{Potapczyk and Przybysz(2020)}]{potapczyk-przybysz-2020-srpols}
Tomasz Potapczyk and Pawel Przybysz. 2020.
\newblock \href {https://doi.org/10.18653/v1/2020.iwslt-1.9} {{SRPOL}{'}s
  system for the {IWSLT} 2020 end-to-end speech translation task}.
\newblock In \emph{Proceedings of the 17th International Conference on Spoken
  Language Translation}, pages 89--94, Online. Association for Computational
  Linguistics.

\bibitem[{Qi et~al.(2020)Qi, Zhang, Zhang, Bolton, and
  Manning}]{qi-etal-2020-stanza}
Peng Qi, Yuhao Zhang, Yuhui Zhang, Jason Bolton, and Christopher~D. Manning.
  2020.
\newblock \href {https://doi.org/10.18653/v1/2020.acl-demos.14} {{S}tanza: A
  python natural language processing toolkit for many human languages}.
\newblock In \emph{Proceedings of the 58th Annual Meeting of the Association
  for Computational Linguistics: System Demonstrations}, pages 101--108,
  Online. Association for Computational Linguistics.

\bibitem[{Sanabria et~al.(2018)Sanabria, Caglayan, Palaskar, Elliott, Barrault,
  Specia, and Metze}]{sanabria18how2}
Ramon Sanabria, Ozan Caglayan, Shruti Palaskar, Desmond Elliott, Lo\"ic
  Barrault, Lucia Specia, and Florian Metze. 2018.
\newblock \href {https://hal.archives-ouvertes.fr/hal-02431947} {{How2: A
  Large-scale Dataset For Multimodal Language Understanding}}.
\newblock In \emph{Proceedings of Visually Grounded Interaction and Language
  (ViGIL)}, Montr{\'e}al, Canada. Neural Information Processing Society
  (NeurIPS).

\bibitem[{Sperber et~al.(2020)Sperber, Setiawan, Gollan, Nallasamy, and
  Paulik}]{sperber-etal-2020-consistent}
Matthias Sperber, Hendra Setiawan, Christian Gollan, Udhyakumar Nallasamy, and
  Matthias Paulik. 2020.
\newblock \href {https://doi.org/10.1162/tacl_a_00340} {Consistent
  transcription and translation of speech}.
\newblock \emph{Transactions of the Association for Computational Linguistics},
  8:695--709.

\bibitem[{Srivastava et~al.(2014)Srivastava, Hinton, Krizhevsky, Sutskever, and
  Salakhutdinov}]{srivastava2014dropout}
Nitish Srivastava, Geoffrey~E Hinton, Alex Krizhevsky, Ilya Sutskever, and
  Ruslan Salakhutdinov. 2014.
\newblock Dropout: a simple way to prevent neural networks from overfitting.
\newblock \emph{Journal of machine learning research}, 15(1):1929--1958.

\bibitem[{Sydorenko(2010)}]{Sydorenko-2012-vocab}
Tetyana Sydorenko. 2010.
\newblock Modality of input and vocabulary acquisition.
\newblock \emph{Lang Learn Technol}, pages 50--73.

\bibitem[{Vaswani et~al.(2017)Vaswani, Shazeer, Parmar, Uszkoreit, Jones,
  Gomez, Kaiser, and Polosukhin}]{vaswani2017attention}
Ashish Vaswani, Noam Shazeer, Niki Parmar, Jakob Uszkoreit, Llion Jones,
  Aidan~N Gomez, {\L}ukasz Kaiser, and Illia Polosukhin. 2017.
\newblock Attention is all you need.
\newblock In \emph{Advances in Neural Information Processing Systems}, pages
  6000--6010.

\bibitem[{Volk et~al.(2010)Volk, Sennrich, Hardmeier, and
  Tidstr{\"o}m}]{Volk-et-al-2010}
Martin Volk, Rico Sennrich, Christian Hardmeier, and Frida Tidstr{\"o}m. 2010.
\newblock {Machine Translation of TV Subtitles for Large Scale Production}.
\newblock In \emph{Proceedings of the Second Joint EM+/CNGL Workshop "Bringing
  MT to the User: Research on Integrating MT in the Translation Industry
  (JEC'10)}, pages 53--62, Denver, CO.

\bibitem[{Weiss et~al.(2017)Weiss, Chorowski, Jaitly, Wu, and
  Chen}]{weiss2017sequence}
Ron~J. Weiss, Jan Chorowski, Navdeep Jaitly, Yonghui Wu, and Zhifeng Chen.
  2017.
\newblock {Sequence-to-Sequence Models Can Directly Translate Foreign Speech}.
\newblock In \emph{Proceedings of Interspeech 2017}, pages 2625--2629,
  Stockholm, Sweden.

\end{thebibliography}

%\appendix

\end{document}